\documentclass[letterpaper, 10 pt, conference]{ieeeconf}  

\IEEEoverridecommandlockouts                              

\let\labelindent\relax

\usepackage[
    style=ieee,
    doi=false,
    isbn=false,
    url=false,
    eprint=false,
    backend=bibtex,
    natbib=true,
    minnames=1,
    maxcitenames=1
    ]{biblatex}

\usepackage{hyperref}
\hypersetup{
    colorlinks=true,
    linkcolor=black,
    urlcolor=blue,
    citecolor=black,
    }
    
\usepackage[shortcuts]{extdash}
\usepackage{graphicx}
\usepackage{mathtools}
\usepackage{dcolumn}
\usepackage{gensymb}
\usepackage{perl_acronyms}
\usepackage{xcolor}
\usepackage{subcaption}
\usepackage{multirow}
\usepackage{amsfonts}
\usepackage{booktabs}
\usepackage{arydshln}
\usepackage{siunitx}
\usepackage{todonotes}
\usepackage{algorithm}
\usepackage{algpseudocode}
\usepackage[inline]{enumitem}
\usepackage{array}
\usepackage{flushend}
\newcolumntype{P}[1]{>{\centering\arraybackslash}p{#1}}

\title{OASIS: Real-Time Opti-Acoustic Sensing for Intervention Systems in Unstructured Environments}

\author{Amy Phung$^{1,2}$, Richard Camilli$^{2}$%
\thanks{This work was supported by the Strategic Environmental Research and Development Program Grant W912HQ24P0024. Amy Phung would like to acknowledge financial support from the National Science Foundation Graduate Research Fellowship (No. 2141064), from the National Aeronautics and Space Administration (NASA) through the FINESST program (No. 80NSSC23K1391)}
\thanks{$^{1}$Applied Ocean Physics and Engineering, Woods Hole Oceanographic Institution, Deep Submergence Laboratory, Woods Hole, MA, USA}
\thanks{$^{2}$Massachusetts Institute of Technology, Cambridge, MA, USA
        {\tt\footnotesize aphung@mit.edu}}%
\thanks{Supplemental Video: \url{https://youtu.be/8Vq9NFSO1cE}}
\thanks{The code, dataset, and visualization tools used to generate the figures in this paper are available online at \url{https://oasis-iros.github.io/}}
}

\begin{document}

\maketitle

\begin{abstract}
\label{sec:abstract}
High resolution underwater 3D scene reconstruction is crucial for various applications, including construction, infrastructure maintenance, monitoring, exploration, and scientific investigation. Prior work has leveraged the complementary sensing modalities of imaging sonars and optical cameras for opti-acoustic 3D scene reconstruction, demonstrating improved results over methods which rely solely on either sensor. However, while most existing approaches focus on offline reconstruction, real-time spatial awareness is essential for both autonomous and piloted underwater vehicle operations. This paper presents OASIS, an opti-acoustic fusion method that integrates data from optical images with voxel carving techniques to achieve real-time 3D reconstruction unstructured underwater workspaces. Our approach utilizes an ``eye-in-hand'' configuration, which leverages the dexterity of robotic manipulator arms to capture  multiple workspace views across a short baseline. We validate OASIS through tank-based experiments and present qualitative and quantitative results that highlight its utility for underwater manipulation tasks.
\end{abstract}

\section{Introduction}
\label{sec:introduction}
Robust perception is essential for safe and efficient completion of subsea intervention tasks in both structured and unstructured environments. Applications range from scientific exploration~\cite{billings2022hybrid} to underwater infrastructure construction~\cite{rizzini2017integration} and maintenance ~\cite{zhang2024advanced}, disaster response~\cite{camilli2012acoustic}, and remediation of hazardous materials~\cite{chizeck2014haptic}. For autonomous and teleoperated robotic vehicles,  intervention operations require sufficient perception for semantic scene understanding and spatial awareness to ensure safe navigation and obstacle avoidance. Existing vehicles predominantly rely on optical cameras for perception~\cite{cai2023autonomous, cieslak2015autonomous, simetti2020autonomous}, which provide human-interpretable visual information but lack robustness in geometric 3D reconstruction, particularly in unstructured environments. Conversely, acoustic sensors offer reliable range data but can be difficult to interpret due to factors such as resolution, field of view, and unintuitive epipolar geometry. 

Prior work has investigated opti-acoustic fusion, which combines the complementary advantages of both optical and acoustic sensors: optical sensors provide high-resolution color, texture, and semantic information, while acoustic sensors offer spatial geometry and depth data~\cite{singh2024opti, qadri2024aoneus, babaee20133}. 
However, creating a dense 3D reconstruction from these data sources is challenging due to significant differences in these sensing modalities' resolutions, dynamic ranges, and fields of view. 
While prior work has demonstrated offline opti-acoustic reconstruction methods with improved results compared to using either sensor alone~\cite{johnson2009towards, bingham2010robotic, qadri2024aoneus}, there remains a critical need for a real-time process that can provide feedback to human pilots or autonomous systems during ongoing vehicle operations. 

We propose OASIS, a real-time opti-acoustic fusion method that integrates optical imagery with voxel carving techniques to enable 3D reconstruction in unstructured underwater environments. Our approach employs an ``eye-in-hand'' configuration, leveraging the dexterity of robotic manipulator arms to maximize the diversity of views across a short baseline. Experimental validation through tank-based tests demonstrates our method's effectiveness in enhancing real-time underwater perception in an underwater manipulation context. To summarize, the key contributions of this work are as follows:

\begin{itemize}
    \item A novel opti-acoustic fusion method capable of achieving real-time performance in both structured and unstructured environments 
    \item A pre-processing method for deringing and normalizing sonar data
    \item An ``eye-in-hand'' method which employs a trajectory that moves the arm minimally from its stored position to ensure safe and efficient exploration of the workspace with few assumptions about the environment's geometry
    \item A quantitatively accurate 3D reconstruction of features within the manipulator work space that is intuitively understandable for humans
    \item Evaluation of the method's performance and limitations using a tank-based dataset 
\end{itemize}

\section{Related Work}
\label{sec:related}
Numerous prior works have developed sonar-based methods for underwater 3D scene reconstruction, which utilize the forward-looking imaging sonars commonly found on underwater vehicles. These methods can be categorized as feature-based~\cite{wang2019underwater}, contour-based~\cite{aykin2013forward}, model-based~\cite{debortoli2019elevatenet, ferretti2023acoustic}, and volumetric techniques~\cite{arnold2022spatial}. Feature-based approaches have been successfully applied to SLAM but lack the ability to model the environment's geometry as a dense 3D reconstruction. Contour and model-based methods typically require prior knowledge or make assumptions about scene geometry (e.g., flat seafloors and monotonic surfaces~\cite{aykin2013forward}), vehicle motion (e.g., rectilinear transits with passively stable platforms~\cite{ferretti2023acoustic}),
or material properties~\cite{debortoli2019elevatenet}. Volumetric techniques reconstruct the environment based on intersecting regions among a collection of sonar images~\cite{aykin2013feature, aykin20153, guerneve2018three, 10753694}. Unlike other approaches, volumetric methods do not impose constraints on environmental structure or vehicle motion. Instead, their reconstruction quality improves with data collected from diverse viewpoints. However, relying solely on acoustic data can limit the interpretability of these reconstructions, making it challenging to accurately identify objects and understand the environment without prior knowledge or additional data from other sensors.

Numerous prior works have investigated the use of opti-acoustic perception methods to improve upon the quality of sonar-only 3D reconstructions. Early works focused on a geometric approach which leverage the sensors’ epipolar geometry~\cite{negahdaripour2009opti, negahdaripour2007epipolar}. While this approach has been widely used for extrinsic calibration of the sensors with known targets~\cite{negahdaripour2009opti, chemisky2021portable, hurtos2010integration}, it requires dense feature correspondences between the optical and acoustic images. However, establishing these correspondences remains a challenge due to differences in the sensors’ fields of view and object appearance with the two sensing modalities. A contour-based approach proposed in~\cite{babaee20153} improves the robustness of earlier feature-based methods but requires a 180$^{\circ}$ rotation of viewing angle around to object for complete reconstruction. However, since this method focuses on object edges and discards information contained in the remaining pixels in the image, a large number of images are required for high-resolution reconstruction. Additionally, this method is optimized for reconstructing discrete objects rather than capturing the overall scene geometry.

The AoNeuS method represents the current state-of-the-art in accurate high-resolution 3D surfaces from opti-acoustic measurements captured over heavily-restricted baselines, where acquiring full 360$^{\circ}$ views of objects may not be possible~\cite{qadri2024aoneus}. This method directly builds the neural surface reconstruction method NeuS~\cite{wang2021neus}, which has gained popularity for its efficiency in representing 3D objects or scenes from optical imagery. 
However, one drawback of NeuS and other neural-based methods is their computational complexity, precluding their utility in real-time applications. NeuS encodes scene geometry using a Signed Distance Function (SDF), which is optimized by sampling along cast rays from the sensor. This encoding requires significant computational resources to process and can take hours to optimize, with typical rendering times of $\sim$5 minutes per frame depending on the resolution and dataset size~\cite{wang2021neus}. 

For optical image datasets, the introduction of 3D Gaussian splatting (3DGS) techniques has addressed some of the computational challenges associated with neural-based methods by eliminating the need to sample along projected rays~\cite{kerbl20233d}. 3DGS has been used for scene reconstruction with both in-air and underwater optical datasets, with modifications to account for the backscatter effects present in underwater imagery~\cite{yang2024seasplat, li2024watersplatting}. Optimization times with 3DGS are typically less than an hour, and the method can achieve real-time rendering performance while maintaining a reconstruction quality comparable to neural-based methods. The RTG-SLAM method~\cite{peng2024rtg} extends 3DGS by utilizing depth imagery to better initialize the Gaussian representation, thus enabling both the optimization and rendering processes to occur in real time. However, 3DGS has not yet been applied to opti-acoustic datasets. 

A comparison of different 3D reconstruction methods is provided in Table~\ref{tab:compare}.

\begin{table}[]
\caption{Comparison of different reconstruction methods}
\label{tab:compare}
\begin{tabular}{c|c|c|c|c}
\hline
Method & \begin{tabular}[c]{@{}c@{}}Optimization \\ Time\end{tabular} & \begin{tabular}[c]{@{}c@{}}Rendering\\ Time\end{tabular} & Incremental? & \begin{tabular}[c]{@{}c@{}}Input \\ Sensors\end{tabular} \\ \hline
COLMAP & \textgreater 1 Hr & \begin{tabular}[c]{@{}c@{}}Real-\\ Time\end{tabular} & N & Optical \\
AoNeuS & 30 min & Minutes & N & \begin{tabular}[c]{@{}c@{}}Optical \&\\ Acoustic\end{tabular} \\
3DGS & 30 min & \begin{tabular}[c]{@{}c@{}}Real-\\ Time\end{tabular} & Y* & Optical \\
\begin{tabular}[c]{@{}c@{}}RTG-\\ SLAM\end{tabular} & Real-time & \begin{tabular}[c]{@{}c@{}}Real-\\ Time\end{tabular} & Y & \begin{tabular}[c]{@{}c@{}}Optical \&\\ Depth\end{tabular} \\
OASIS & \begin{tabular}[c]{@{}c@{}}Real-time \\ (18 Hz)\end{tabular} & \begin{tabular}[c]{@{}c@{}}Real-\\ Time\end{tabular} & Y & \begin{tabular}[c]{@{}c@{}}Optical \& \\ Acoustic\end{tabular} \\ \hline
\end{tabular}
\vspace{0.1pt}
\\
\footnotesize *Possible in certain implementations
\end{table}

\section{Method}
\label{sec:method}
This method extends prior work on sonar-based 3D reconstruction to include optical data in the reconstruction. It extends the min-max approach~\cite{10753694} for reasoning over scene geometry using sonar data, which is then used to render corresponding depth images for each optical image. These images are then projected onto the 3D reconstruction to provide semantically meaningful color information on objects in the workspace. The overall reconstruction pipeline can be described as follows:
\begin{enumerate}
    \item Record sonar data across a minimal ``sweep'' trajectory to map obstacle-free regions in the workspace
    \item Compute initial occupancy grid from sonar data
    \item Use occupancy grid to obtain close-up views with an optical camera
    \item Fuse optical images with occupancy grid 
\end{enumerate}
Each of these key steps in the reconstruction process are discussed in the following subsections. 

\subsection{Preliminary Workspace Mapping}
\label{sec:mapping}

\begin{figure}[h]
  \includegraphics[width=\linewidth]{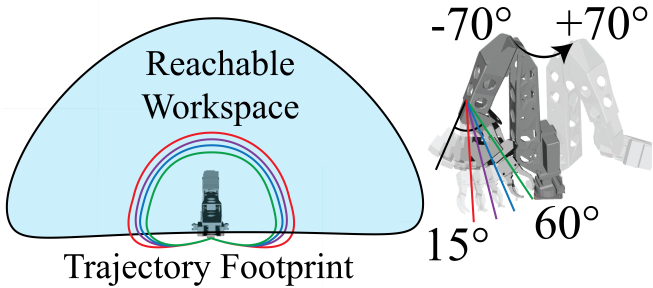}
  \caption{The proposed ``sweep'' trajectory records a series of intersecting imaging sonar images which are used to reconstruct the workspace geometry while requiring minimal movement beyond the manipulator's stowed position.}
  \label{fig:trajectory}
\end{figure}

We propose an eye-in-hand mapping approach, which uses a wrist-mounted optical camera and imaging sonar for workspace perception. Before conducting intervention tasks, it is essential to map the workspace to identify obstacles. Therefore, in an unmapped workspace, imaging sonar data is collected using a low-profile ``sweep'' trajectory which moves the arm minimally from its stowed position to generate a preliminary map while minimizing collision risk. 
Elevation angle ambiguity is minimized in the sonar reconstruction by setting the manipulator's wrist yaw to its limit (+$50^\circ$), with the wrist pitch and elbow angles set such that the sonar is pitched downwards ($60^\circ$ and $22^\circ$, respectively). In this configuration, the arm's shoulder yaw joint is actuated from $-70^\circ$ to $70^\circ$, which provides a sweep of data across the workspace. A corresponding sweep is recorded by setting the wrist yaw set to its other limit (-$50^\circ$) and actuating the shoulder yaw joint in the other direction (from $70^\circ$ to $-70^\circ$). This back-and-forth sweeping motion enables sonar observation at perpendicular roll angles and is repeated for a series of wrist pitch angles ($45^\circ$, $30^\circ$, and $15^\circ$), which provides vertical coverage of the workspace. This trajectory is illustrated in Figure~\ref{fig:trajectory}.

\subsection{Acoustic 3D-Reconstruction}
\label{sec:sonar-mapping}

\begin{figure*}[ht!]
    \centering
  \includegraphics[width=\linewidth]{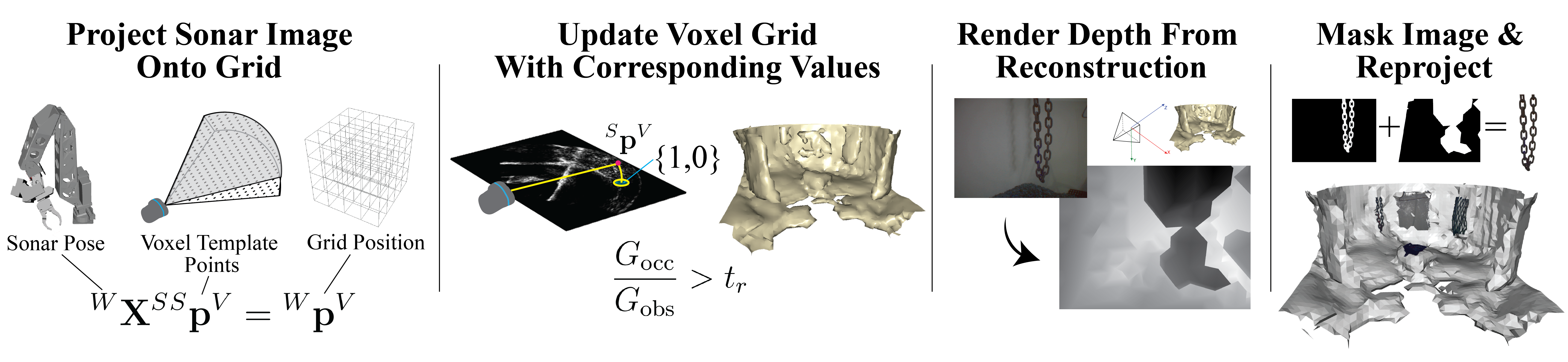}
  \caption{Reconstruction process overview}
  \label{fig:pipeline}
\end{figure*}

\begin{figure}[]
    \centering
  \includegraphics[width=\linewidth]{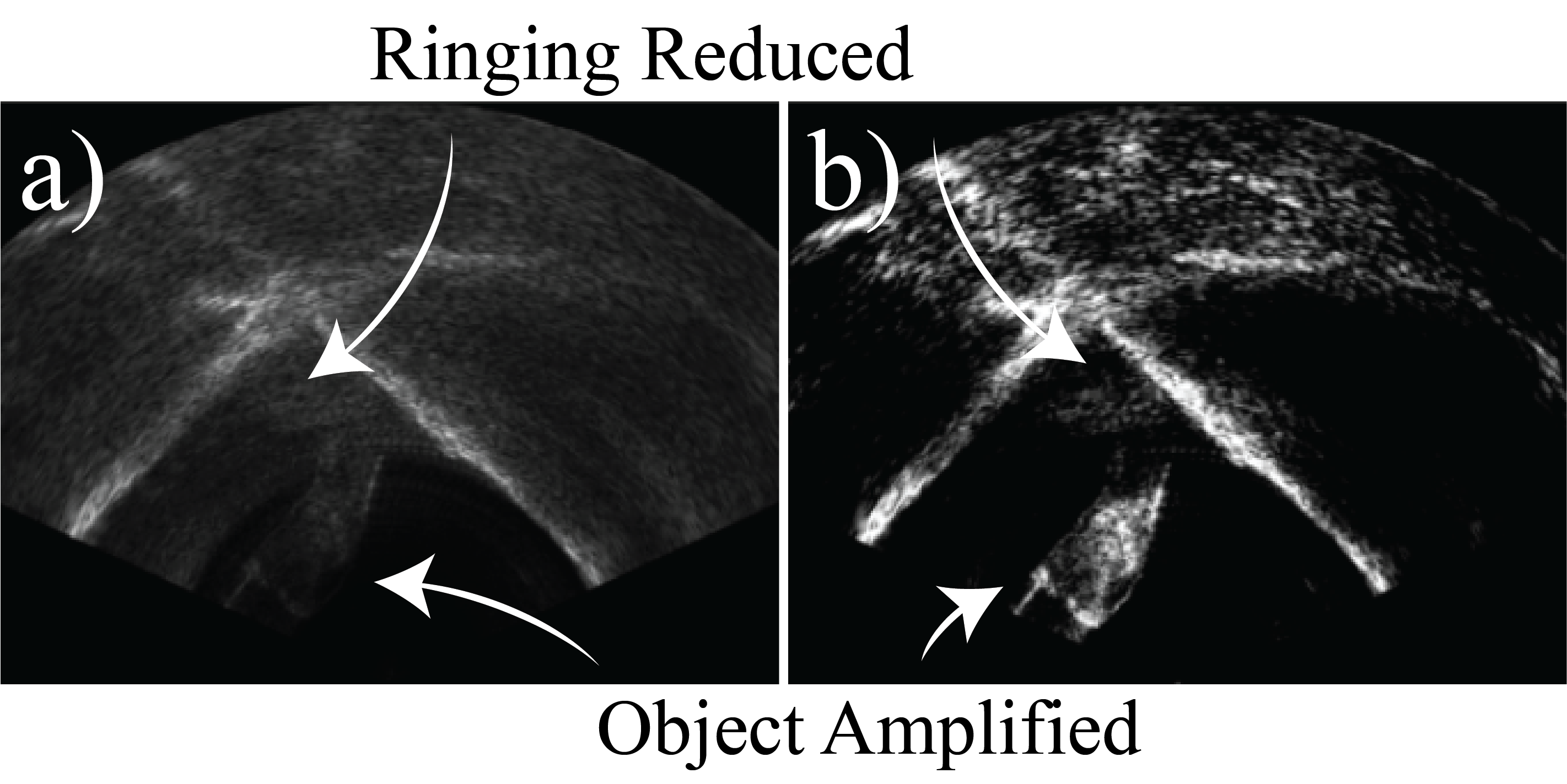}
  \caption{The pre-processing step is applied to the raw data (a) to produce a binary occupancy map (b) with less ringing}\label{fig:preprocessing}
\end{figure}

Elevation angle ambiguity, which is caused by the wide vertical aperture of imaging sonar beams, poses a challenge for 3D reconstruction. Prior work by ~\cite{xu2011method} and others use acoustic concentrator lensing to reduce this ambiguity. However, the approach presents a trade-off of decreased field-of-view, which increases acquisition time requirements and decreases its utility for real-time perception processes. 
We instead address the elevation angle ambiguity by employing a volumetric 3D reconstruction technique based on~\cite{10753694}, which evaluates intersecting regions across multiple sonar images to construct a voxel grid representation of the scene (Alg. 1). This method is particularly well-suited for an ``eye-in-hand'' approach since it leverages the manipulator’s ability to collect data across a wide variety of poses, which in turn helps to resolve the elevation angle ambiguity present in the data.

\begin{algorithm}
\caption{Acoustic 3D reconstruction}
\begin{algorithmic}[1]
\Statex \textbf{Input} 
\Statex \hspace{9pt} $D$: Set of sonar data frames
\Statex \textbf{Output} 
\Statex \hspace{9pt} $V$: Voxel grid
\State $G_\text{occ}$ Initialize count of occupancy voxel grid across manipulator reachable workspace, with voxel resolution
\State $G_\text{obs}$ Initialize count of observations
\State $T$ Construct voxel template
\For{each new sonar data frame $S_c$ in $D$}
    \State $\bar{S}_c$ = \textsc{Preprocess}($S_c$) 
    \State Project values of $\bar{S}_c$ into $T$
    \For {each voxel ${}^{S}_{}\mathbf{p}^{V}_{}$ in $T$}  
        \If{$G_\text{occ}({}^{S}_{}\mathbf{p}^{V}_{}) /G_\text{obs}({}^{S}_{}\mathbf{p}^{V}_{}) >t_r$}
            \State $T({}^{S}_{}\mathbf{p}^{V}_{}) = 1$
        \Else 
            \State $T({}^{S}_{}\mathbf{p}^{V}_{}) = 0$
        \EndIf
    \EndFor
    \State ${}^{W}_{}\mathbf{X}^{S}_{}$ = \textsc{LookupSonarPose}($S_c$)
    \State ${}^{W}_{}\mathbf{p}^{V}_{} = {}^{W}_{}\mathbf{X}^{S}_{}
    {}^{S}_{}\mathbf{p}^{V}_{}$
    \State $V({}^{W}_{}\mathbf{p}^{V}_{}) = 
    T({}^{W}_{}\mathbf{p}^{V}_{})$
\EndFor
\end{algorithmic}
\end{algorithm}

\textbf{Step 1: Initialization}
A voxel grid, $V$, and two corresponding 3D meshgrids of the same size, are initialized with the dimensions of the grids fully containing the manipulator workspace. The two meshgrids $G_\text{obs}$ and $G_\text{occ}$ track the number of times each voxel has been observed and marked as occupied, respectively. 

A ``voxel template''~\cite{10753694} is then constructed based on the voxel grid resolution, which contains the minimum number of points needed to sufficiently represent the data's vertical and horizontal fields-of-view (FOV). Rather than projecting each individual beam and range bin in the data into the global frame during the map update step, this template significantly reduces the number of projections required, thus enabling the map update to occur in real-time.

\textbf{Step 2: Preprocessing}
To improve the performance of the reconstruction process, sonar data is only processed if the position of the new data differs by more than 1 cm from the position of the previously processed data. A pre-processing step is applied to the data to remove ringing artifacts and normalize the intensity values across different ranges. During pre-processing, the input data is converted from its native polar coordinates, where the rows indicate ranges and the columns indicate bearings, into cartesian coordinates.   
A standard-deviation-based approach is then applied to convert the 8-bit sonar ping return intensity information to a binary occupancy map. The first 10 range bins across all beams ($\sim5$cm) are assumed to be empty, and are used to characterize the mean $\mu_\text{bg}$ and standard deviation $\sigma_\text{bg}$ of the background noise. For non-empty range bins in the remainder of the data (i.e., where a beam's range bin contains an intensity value greater than $\mu_\text{bg} + 2\sigma_\text{bg}$), we construct a rolling window $W$ that sequentially iterates through the sonar return data's entire bin range. Each iteration contains an ensemble of each of the 512 beam's ping return intensity values for a given range bin. This range ensemble is used to compute the mean $\mu_\text{W}$ and standard deviation $\sigma_\text{W}$ for each range bin. Each beam's return intensity at a given range bin is then considered occupied if its intensity value is greater than one standard deviation from the mean ($\mu_\text{W} + \sigma_\text{W}$). This statistical process of normalizing return intensity is used to suppress ring generated by high-intensity acoustic returns, which can lead to false-positive artifacts in the 3D reconstruction. 

Results from this pre-processing method are illustrated in Figure~\ref{fig:preprocessing}. 

\begin{algorithm}
\caption{Sonar Data Preprocessing}
\begin{algorithmic}[1]
\Statex \textbf{Input} 
\Statex \hspace{9pt} $S_p$: Sonar intensity data (in polar coordinates)
\Statex \hspace{9pt} $w$: Range window
\Statex \textbf{Output} 
\Statex \hspace{9pt} $\bar{S}_c$: Binary occupancy map (cartesian)
\State $\mu_{\text{bg}}$ = \textsc{Mean}($S_c$[:10, :]) 
\State $\sigma_{\text{bg}}$ = \textsc{Std}($S_c$[:10, :]) 
\For{each row $r$ in $S_p$}
    \If{\textsc{Max}($S_p$[r, :]) $\geq$ $\mu_\text{bg} + 2\sigma_\text{bg}$}
        \State $W = S_p[r-w:r+w, :]$ 
        \State $\mu_{\text{w}}$ = \textsc{Mean}($W$)
        \State $\sigma_{\text{w}}$ = \textsc{Std}($W$)
        \For{each pixel $p$ in $r$}
            \If{$p > \mu_\text{w} + \sigma_{\text{w}}$}
                \State $\bar{S}_p[r, p] = 1$
            \Else 
                \State $\bar{S}_p[r, p] = 0$
            \EndIf
        \EndFor
    
    \Else
        \State $\bar{S}_p[r, :] = 0$
    \EndIf
\EndFor
\State $\bar{S}_c$ = \textsc{ConvertToCartesian}($\bar{S}_p$) 
\end{algorithmic}
\end{algorithm}
\pagebreak
\textbf{Step 3: Projection to Global Frame}
The voxel template from Step 1 is projected onto the binary occupancy image produced in Step 2. Each voxel in the template is marked as occupied if any of the pixels contained within it indicate occupancy; otherwise, it remains unoccupied. This voxel template is then projected into the world frame using the sonar's estimated pose based on the manipulator arm's forward kinematics model and joint angle sensors.

\textbf{Step 4: Voxel Grid Update}
For each voxel in the world frame-projected voxel template, we increment the corresponding value in the meshgrid $G_\text{obs}$ to indicate a new observation. If the voxel indicates an occupancy, the corresponding value in $G_\text{occ}$ is also incremented. The ratio between these values is compared to a threshold $t_r$, which is determined empirically based on the data's false negative rate. Voxels with a ratio greater than $t_r$ are then marked as occupied, and the remaining voxels are marked as unoccupied.

A variety of factors, such as speckle noise, acoustic shadows, and oblique incidence angles can cause false negative artifacts. Sensitivity analysis indicates that if $t_r$ is set to 1, this will result in gaps (i.e., holidays) in the map. At the other extreme, if $t_r = 0$, the resulting map will be gap-free but elevation angle ambiguities will cause an increased rate of false-positive regions in the map.

\subsection{Opti-Acoustic Fusion}

Using this geometric representation of the acoustic information, a trajectory can be selected to record optical frames. Typically, due to backscatter and light attenuation, optical imagery requires closer stand-off distances than sonar imagery. In isolation, the sonar reconstruction lacks visual detail, making object recognition by humans difficult without prior knowledge of the workspace. However, the acoustic reconstruction is adequate for locating objects in the scene, and can be used to maneuver the arm around obstacles to obtain closer stand-off distances while recording optical frames.

To fuse the optical data with the sonar data, we project the optical data onto the voxel grid and render it as an overlay. The voxel grid is first converted to a mesh using Open3D and then smoothed using the marching cubes algorithm~\cite{lorensen1998marching, pmneila_pymcubes_2025}. Using the optical camera's intrinsics (extracted from the calibration process) and extrinsics (derived from the manipulator arm's joint angle sensors and forward kinematics model), a virtual camera is constructed and used to render a corresponding depth image using the meshed reconstruction for each recorded optical image. A background mask is constructed for the optical and depth image using Rembg~\cite{gatis_rembg} with the ISNet model~\cite{qin2022} and the watershed algorithm~\cite{87344}, respectively. The intersection of these masks is applied to both images to segment the object. Using the camera pose, intrinsics, and depth image, the segmented pixels are projected into the world frame, and rendered alongside the meshed voxel grid. This process is illustrated in Figure~\ref{fig:pipeline}.

\section{Evaluation}
\label{sec:results}
\begin{figure}[h]
  \centering
\includegraphics[width=0.5\linewidth]{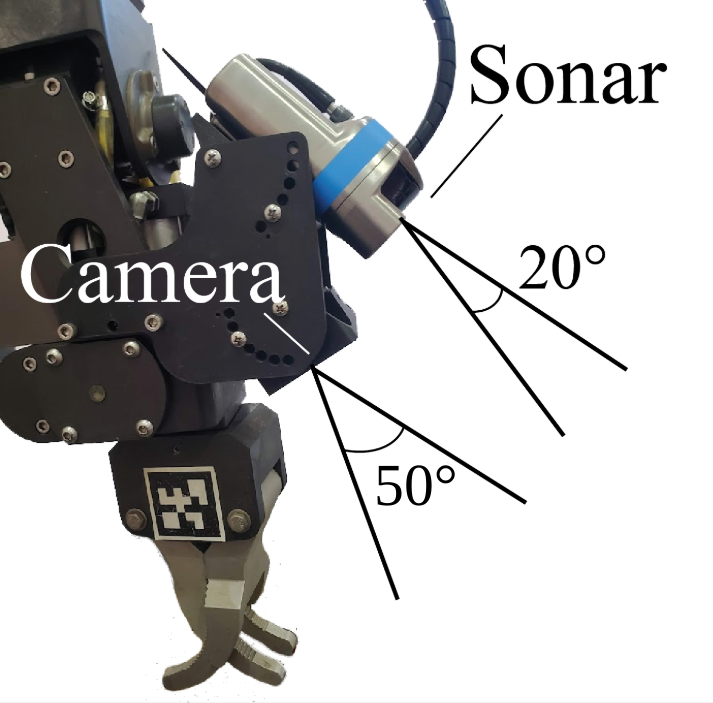}
  \caption{Experiments are conducted with an opti-acoustic eye-in-hand approach. The top half of the camera field of view overlaps with the sonar's in the manipulator workspace}
  \label{fig:setup}
\end{figure}

To validate our approach, we conduct tank-based experiments with an Oculus M1200d multibeam imaging sonar (Blueprint Subsea) operating at 1.2MHz and a stellarHD Global Shutter Frame-Sync Subsea ROV/AUV USB Machine Vision Camera (DeepWater Exploration). These sensors are mounted on the wrist of a fixed-base seven degree-of-freedom hydraulic manipulator arm (Kraft TeleRobotics) (pictured in Figure~\ref{fig:setup}). 

The sonar has 512 horizontal beams spanning a 130$^\circ$ horizontal field-of-view (FOV), and has a 20 degree vertical FOV. Data is recorded at the sensor's maximum framerate of 10 FPS and stored as an 512x398 image, where each pixel contains an 8-bit intensity value for the corresponding horizontal beam and range bin. During the experiments, the sonar was configured with a 2 m maximum range and a gain setting of 10.  The camera uses a 1/2.9” Omnivision OmniPixel 3-GS™ CMOS sensor and has a maximum resolution of 1600x1200. The camera was operated at 10 FPS during the experiments.

\begin{figure*}[ht!]
  \includegraphics[width=\textwidth]{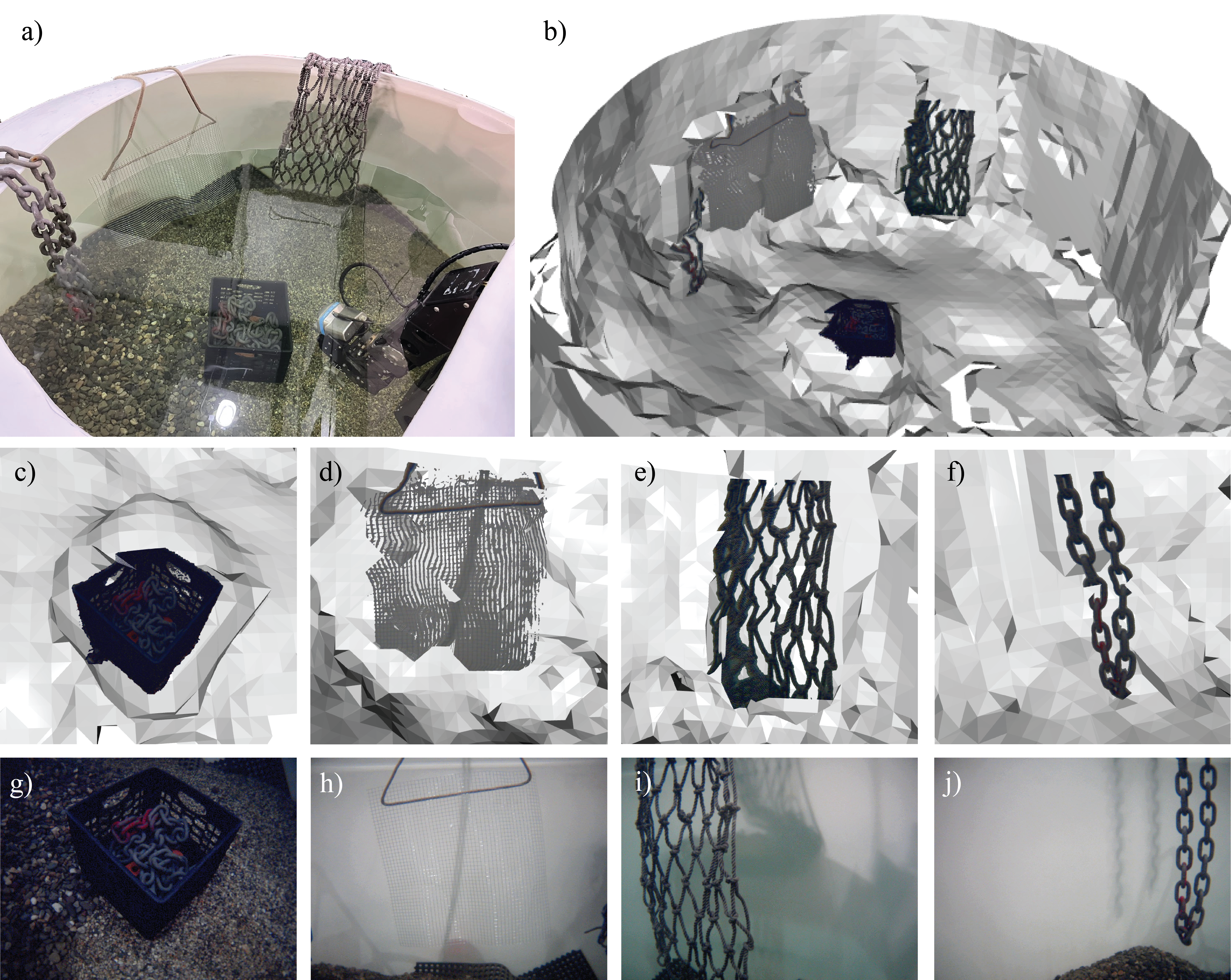}
  \caption{Comparison of reconstruction results for the tank (a,b), milk crate (c,g), metal mesh (d,h), cargo net (e,i), and chain (f,j).}
  \label{fig:compare}
\end{figure*}

Dataset collection was conducted in a 2.1 m diameter, 1.5 m deep tank, with objects chosen for their varying materials and geometries. These include a plastic milk crate, wire mesh, cargo net, and chain (Figure~\ref{fig:compare}a). The ``sweep'' trajectory described in Section~\ref{sec:mapping} was used to record sonar data at its maximum framerate, and took approximately 90 seconds to complete. Afterwards, the arm was manually operated to capture close-up optical images of each of object. A side-by-side comparison of the reconstruction results using this tank setup is shown in Figure~\ref{fig:compare}. The reconstruction was generated with a voxel grid resolution of 0.05 m.

\begin{figure}[]
  \includegraphics[width=\linewidth]{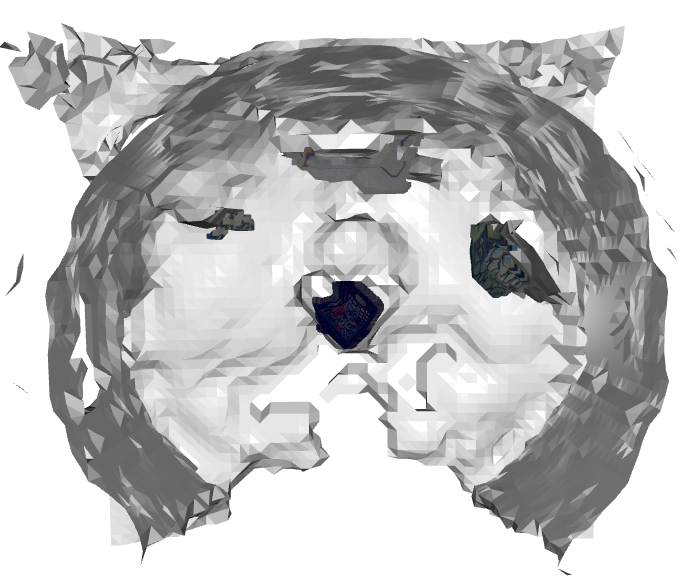}
  \caption{Top view of reconstructed results. The tank's back wall is fully reconstructed despite partial occlusions from objects in the tank}
  \label{fig:top}
\end{figure}

\begin{table}[h]
\centering
\caption{Processing times and equivalent framerates per sonar frame based on voxel size}
\begin{tabular}{ccc}
\hline
\begin{tabular}[c]{@{}c@{}}Voxel Size\\ (m)\end{tabular} & \begin{tabular}[c]{@{}c@{}}Time Per Frame \\ (s)\end{tabular} & \begin{tabular}[c]{@{}c@{}}Framerate\\ (FPS)\end{tabular} \\ \hline
0.05                                                     & 0.052                                                         & 18.6                                                      \\
0.04                                                     & 0.116                                                         & 8.6                                                       \\
0.03                                                     & 0.195                                                         & 5.1                                                       \\
0.02                                                     & 0.501                                                         & 3.4                                                       \\
0.01                                                     & 3.516                                                         & 0.28                                                      \\ \hline
\end{tabular}
\label{tab:time}
\end{table}

\begin{table}[H]
\centering
\caption{Optical and acoustic reconstruction accuracy, in centimeters}
\begin{tabular}{cccccc}
\hline
\multicolumn{1}{c|}{\multirow{2}{*}{Object}} & \multicolumn{1}{c|}{\multirow{2}{*}{\begin{tabular}[c]{@{}c@{}}Ground\\ Truth\end{tabular}}} & \multicolumn{2}{c|}{Acoustic} & \multicolumn{2}{c}{Optical} \\ \cline{3-6} 
\multicolumn{1}{c|}{} & \multicolumn{1}{c|}{} & \multicolumn{1}{l|}{Result} & \multicolumn{1}{l|}{Error} & \multicolumn{1}{l|}{Result} & \multicolumn{1}{l}{Error} \\ \hline
\multicolumn{1}{c|}{Tank} & \multicolumn{1}{c|}{\begin{tabular}[c]{@{}c@{}}220.5 \\ (0.4)\end{tabular}} & \multicolumn{1}{c|}{\begin{tabular}[c]{@{}c@{}}215.3 \\ (1.0)\end{tabular}} & \multicolumn{1}{c|}{5.2} & \multicolumn{1}{c|}{-} & - \\
\multicolumn{1}{c|}{Milk Crate} & \multicolumn{1}{c|}{\begin{tabular}[c]{@{}c@{}}32.9 \\ (0.1)\end{tabular}} & \multicolumn{1}{c|}{\begin{tabular}[c]{@{}c@{}}26.3 \\ (0.9)\end{tabular}} & \multicolumn{1}{c|}{6.6} & \multicolumn{1}{c|}{\begin{tabular}[c]{@{}c@{}}22.3\\ (1.4)\end{tabular}} & 10.6 \\
\multicolumn{1}{c|}{\begin{tabular}[c]{@{}c@{}}Mesh\\ (full width)\end{tabular}} & \multicolumn{1}{c|}{\begin{tabular}[c]{@{}c@{}}58.5 \\ (2.1)\end{tabular}} & \multicolumn{1}{c|}{\begin{tabular}[c]{@{}c@{}}53.4 \\ (0.2)\end{tabular}} & \multicolumn{1}{c|}{5.1} & \multicolumn{1}{c|}{-} & - \\
\multicolumn{1}{c|}{\begin{tabular}[c]{@{}c@{}}Mesh\\ (10 cells)\end{tabular}} & \multicolumn{1}{c|}{\begin{tabular}[c]{@{}c@{}}12.7\\ (0.1)\end{tabular}} & \multicolumn{1}{c|}{-} & \multicolumn{1}{c|}{-} & \multicolumn{1}{c|}{\begin{tabular}[c]{@{}c@{}}11.5\\ (0.6)\end{tabular}} & 1.2 \\
\multicolumn{1}{c|}{Chain} & \multicolumn{1}{c|}{\begin{tabular}[c]{@{}c@{}}42.0\\ (0.2)\end{tabular}} & \multicolumn{1}{c|}{-} & \multicolumn{1}{c|}{-} & \multicolumn{1}{c|}{\begin{tabular}[c]{@{}c@{}}33\\ (0.6)\end{tabular}} & 9.0 \\ \hline
\multicolumn{6}{c}{\begin{tabular}[c]{@{}c@{}}*Measurement variance (due to object deformation) \\ indicated in parentheses.\end{tabular}}
\end{tabular}
\label{tab:accuracy}
\end{table}

The method is able to reconstruct the submerged objects as well as the tank wall, as shown in the top-down view of the reconstruction result in Figure~\ref{fig:top}. This indicates that the sonar is capable of capturing information about the workspace geometry despite the partial occlusions and acoustic reverberation generated by the circular tank wall, even when the optical visibility is obstructed. 

With the 0.05 m voxel grid resolution used in these experiments, it took approximately 0.052 seconds to process each new sonar frame. This translates to a theoretical maximum framerate of 18 FPS, thus exceeding sonar's maximum framerate capabilities and indicating real-time performance. For higher voxel grid resolutions, the processing time increases cubically, as the number of points in the voxel template grows at a rate of $O^3$ with respect to the resolution. Processing times and corresponding framerates for various voxel grid resolutions are listed in Table~\ref{tab:time}.

To evaluate the accuracy of the reconstruction, the dimensions of the object based on ground-truth measurements and the opti-acoustic results are reported in Table~\ref{tab:accuracy}. Dimensions which cannot be derived from the data are omitted. Individual wire mesh cells and chain links cannot be identified in the acoustic reconstruction due to the voxel grid resolution. The cargo net's dimensions are not consistent due to its deformability, and thus its measurement accuracy is not reported.

\section{Discussion}
\label{sec:Discussion}
Optical sensors are sensitive to environmental factors such as turbidity and lighting, which makes them less reliable in underwater environments. In deep-sea conditions, the lack of ambient light typically necessitates longer camera integration times, which in turn requires slower movements during data collection to avoid motion blur. Artificial lighting is also required in these environments, which require significantly more power to operate than the sonar. Since lighting conditions are often uneven, cameras are carefully positioned to optimize the use of the available light. For these reasons, cameras cannot be used as quickly or as effectively as sonars can to obtain a broad dataset using a sweeping trajectory.

In an operational context, the acoustic 3D reconstruction could be used to improve the safety and efficiency of intervention tasks with autonomous and teleoperated robotic vehicles. If applied to conventional subsea intervention platforms, the reconstruction could be used to inform the operator which regions are safe for manipulation in real-time, and enable them to avoid collisions or entanglement when obtaining close-up views with the optical camera. Although objects are not identifiable from the acoustic 3D reconstruction alone, the projected optical camera data provides semantically meaningful information that enables the operator to discern which objects are present in the environment. The low computational and power requirements of the method also extends its capabilities to submersible vehicles which carry power onboard (e.g., Nereid-UI~\cite{bowen2014design}, Girona 500 AUV~\cite{ribas2011girona}) and support its use on future untethered vehicle designs.

A key factor in enabling real-time performance is that sonar data can be decimated by recording the maximum intensity over a group of pixels, thus reducing the computational requirements for processing. While this decimation may increase the false positive rate and cause objects to appear larger, it does not increase the false negative rate, allowing the low-resolution reconstruction to retain sufficient fidelity for collision avoidance during operations.

In its current implementation, this method is limited to small, static workspaces, and relies on accurate pose information from a fixed-base manipulator. To address these limitations, future work will explore object-based tracking for dynamic environments and integrate a SLAM-based approach to support free-floating manipulator platforms. 
Additionally, the method's performance in turbid conditions will be evaluated, and the optical fusion capabilities can be improved by incorporating a range-dependent color correction to enable multi-view image stitching.

\section{Conclusion}
\label{sec:conclusion}
In this paper, we present OASIS, a real-time opti-acoustic fusion method that integrates voxel carving and Gaussian splatting techniques for 3D reconstruction in unstructured underwater environments. By leveraging an ``eye-in-hand'' configuration, OASIS captures diverse workspace views while ensuring safe, efficient exploration with minimal assumptions about the environment’s geometry. Experimental validation demonstrates the effectiveness of our approach for underwater manipulation tasks. Key contributions include a novel method for opti-acoustic fusion and sonar preprocessing, a low-profile ``sweep'' trajectory for preliminary mapping, and quantitative and qualitative results from applying the method and trajectory to a tank-based dataset.

\renewcommand{\bibfont}{\normalfont\small}
\printbibliography

\end{document}